\documentclass{article}

\usepackage{microtype}
\usepackage{graphicx}
\usepackage{subfigure}
\usepackage{booktabs} 

\usepackage{makecell}
\usepackage{multirow}
\usepackage{amsmath}

\usepackage{hyperref}

\usepackage[accepted]{icml2025}

\usepackage{amsmath}
\usepackage{amssymb}
\usepackage{mathtools}
\usepackage{amsthm}

\usepackage[capitalize,noabbrev]{cleveref}

\theoremstyle{plain}

\theoremstyle{definition}

\theoremstyle{remark}

\usepackage[textsize=tiny]{todonotes}

\icmltitlerunning{SciPIP: An LLM-based Scientific Paper Idea Proposer}

\newcommand\ie{\textit{i.e.}}
\newcommand\eg{\textit{e.g.}}
\newcommand\etc{\textit{etc}}

\begin{document}

\twocolumn[
\icmltitle{SciPIP: An LLM-based Scientific Paper Idea Proposer}




\begin{icmlauthorlist}
\icmlauthor{Wenxiao Wang}{zju}
\icmlauthor{Lihui Gu}{zju}
\icmlauthor{Liye Zhang}{zju}
\icmlauthor{Yunxiang Luo}{zju}
\icmlauthor{Yi Dai}{zju}
\icmlauthor{Chen Shen}{alibaba}
\icmlauthor{Liang Xie}{zjut}
\icmlauthor{Binbin Lin}{zju}
\icmlauthor{Xiaofei He}{zju}
\icmlauthor{Jieping Ye}{alibaba}
\end{icmlauthorlist}

\icmlaffiliation{zju}{Zhejiang University}
\icmlaffiliation{alibaba}{Alibaba Cloud Computing}
\icmlaffiliation{zjut}{Zhejiang University of Technology}


\icmlkeywords{Machine Learning, ICML}

\vskip 0.3in
]



\printAffiliationsAndNotice{}  

\begin{abstract}
The rapid advancement of large language models (LLMs) has opened new possibilities for automating the proposal of innovative scientific ideas. This process involves two key phases: literature retrieval and idea generation. However, existing approaches often fall short due to their reliance on keyword-based search tools during the retrieval phase, which neglects crucial semantic information and frequently results in incomplete retrieval outcomes. Similarly, in the idea generation phase, current methodologies tend to depend solely on the internal knowledge of LLMs or metadata from retrieved papers, thereby overlooking significant valuable insights contained within the full texts. To address these limitations, we introduce SciPIP, an innovative framework designed to enhance the LLM-based proposal of scientific ideas through improvements in both literature retrieval and idea generation. Our approach begins with the construction of a comprehensive literature database that supports advanced retrieval based not only on keywords but also on semantics and citation relationships. This is complemented by the introduction of a multi-granularity retrieval algorithm aimed at ensuring more thorough and exhaustive retrieval results. For the idea generation phase, we propose a dual-path framework that effectively integrates both the content of retrieved papers and the extensive internal knowledge of LLMs. This integration significantly boosts the novelty, feasibility, and practical value of proposed ideas. Our experiments, conducted across various domains such as natural language processing and computer vision, demonstrate SciPIP's capability to generate a multitude of innovative and useful ideas. These findings underscore SciPIP's potential as a valuable tool for researchers seeking to advance their fields with groundbreaking concepts.
\end{abstract}

\section{Introduction}

With the rapid expansion of scientific knowledge and the increasing complexity of interdisciplinary research, researchers face significant challenges, including information extraction and knowledge synthesis when generating novel ideas. The advent of large language models (LLMs), such as GPT-4~\citep{DBLP:conf/nips/Ouyang0JAWMZASR22}, LLaMA~\citep{DBLP:journals/corr/abs-2302-13971,DBLP:journals/corr/abs-2307-09288}, Qwen~\citep{DBLP:journals/corr/abs-2309-16609,DBLP:journals/corr/abs-2407-10671}, GLM-4~\citep{DBLP:journals/corr/abs-2406-12793}, DeepSeek V3~\cite{DBLP:journals/corr/abs-2412-19437}, and others, provides promising solutions. These models can process vast amounts of literature at speeds far beyond human capabilities, effectively mitigating concerns about information overload. Moreover, their broad interdisciplinary knowledge enables them to synthesize insights across domains, fostering the generation of innovative ideas.

Recent research has explored the potential of LLMs in assisting researchers with idea generation. Most approaches involve two phases: \textit{literature retrieval} and \textit{idea generation}. Despite their promise, both phases have limitations:
\textit{(1) During the literature retrieval phase,} most existing algorithms~\citep{DBLP:journals/corr/abs-2408-06292,DBLP:journals/corr/abs-2409-14634} rely on keyword-based retrieval techniques, which fail to leverage semantic information inherent in textual data, significantly compromising retrieval thoroughness. While some approaches employ semantic information~\citep{DBLP:journals/corr/abs-2410-09403,DBLP:conf/acl/XuSXFWZ23} through vector-based retrieval, they often encode entire sections (\eg, abstracts) into vectors. However, as each section typically contains multifaceted information, such an approach makes it difficult to capture key points effectively. This, in turn, impacts both encoding quality and retrieval performance.
\textit{(2) During the idea generation phase}, 
existing idea generation approaches~\citep{DBLP:journals/corr/abs-2408-06292,DBLP:journals/corr/abs-2404-07738,DBLP:conf/acl/0005DJH24} mainly rely on the LLM's internal knowledge and do not take full advantage of retrieved papers.

To address these challenges, we propose SciPIP, a novel framework designed to enhance idea proposal. In particular, for a more thorough literature retrieval, we first \textbf{construct a literature database}. Specifically, we collect approximately 78K papers from several top-tier academic conferences of AI field and utilize LLMs to re-summarize each paper into a structured quintuple consisting of keywords, backgrounds, ideas, concise methods, and references. These components are individually preprocessed (\eg, encoded into vectors) and stored in a database to enable more precise and efficient retrieval. Based on our constructed database, we propose a \textbf{multi-granularity retrieval method}, which comprehensively leverages keywords, semantic embeddings and citation relations, enabling a thorough literature retrieval across multiple dimensions.

In the idea generation phase, we propose a \textbf{dual-path idea generation framework}. This approach explicitly separates idea generation into two pathways: one leveraging the LLM’s internal knowledge and the other utilizing retrieved literature. The LLM is then guided to integrate these two pathways, synthesizing valuable insights from both to generate comprehensive and well-informed final ideas.

Extensive experiments are conducted to evaluate both idea proposal and literature retrieval on the Natural Language Processing (NLP) and Computer Vision (CV) domains. Various idea generation tools including SciPIP propose ideas based on a given user's query, and the quality of the proposed ideas are assessed by human experts. Results show that SciPIP significantly outperforms existing idea generation approaches in both CV and NLP domains in terms of novelty, feasibility, clarity, relevance, \etc.

The main contributions of this paper include:
\vspace{-3mm}
\begin{itemize}
    \item We construct a literature database for AI papers, with their key information and embeddings extracted, which can benefit the community for future research.
    \vspace{-2mm}
    \item Based on the literature database, we introduce a multi-granularity literature retrieval algorithm.
    \vspace{-2mm}
    \item We present an LLM-based dual-path idea generation framework, improving the proposed research ideas in terms of the novelty, clarity, feasibility, \etc.
    \vspace{-2mm}
    \item Extensive experiments are conducted to demonstrate the effectiveness of SciPIP.
\end{itemize}

\section{Related Works}

\paragraph{LLM-based Scientific Idea Proposal.} Some previous works~\citep{DBLP:journals/corr/abs-2409-14634,DBLP:journals/corr/abs-2410-09403,DBLP:conf/acl/XuSXFWZ23,DBLP:journals/corr/abs-2410-14255,DBLP:journals/corr/abs-2408-06292,DBLP:journals/corr/abs-2404-07738,DBLP:conf/acl/0005DJH24} have explored how to generate research ideas using large language models. These methods typically involve literature retrieval and idea generation. For instance, AI Scientist~\citep{DBLP:journals/corr/abs-2408-06292} relies on keyword-based search engines such as Semantic Scholar for literature retrieval, which fail to leverage the semantic information of the documents. In contrast, \citep{DBLP:journals/corr/abs-2410-09403} employs a FAISS~\citep{johnson2019billion} vector search database, where each vector is encoded from full titles, abstracts, and other metadata. Some approaches such as~\citep{DBLP:journals/corr/abs-2404-07738} also utilize the citation lists of papers to select relevant references.
In the idea generation phase, \citep{DBLP:journals/corr/abs-2408-06292} relies on the internal knowledge of the model to propose ideas. Additionally, AI Scientist refines repetitive ideas through prompt engineering after performing a post-generation literature search, aiming to enhance the novelty of the final ideas. ResearchAgent~\citep{DBLP:journals/corr/abs-2404-07738}, on the other hand, focuses on identifying problems in the retrieved papers and suggesting corresponding solutions.
However, these algorithms primarily utilize only the abstracts of the retrieved papers, often resulting in a lack of necessary details in their methods. In SciPIP, we re-summarize each paper into
a structured quintuple.

\vspace{-2mm}
\paragraph{Retrieval Augmented Generation.} 
Many retrieval-augmented generation (RAG) algorithms have been proposed to enhance performance on various retrieval-required downstream tasks. Typical RAG methods involve three primary steps: indexing, retrieval, and generation~\cite{DBLP:journals/corr/abs-2312-10997}. 
Several mature RAG algorithms have demonstrated effectiveness across diverse applications~\cite{DBLP:journals/corr/abs-2408-17072,DBLP:conf/www/PengLJWOZX0C24,DBLP:conf/acl/GaoMLC23,DBLP:conf/iclr/0002IWXJ000023,DBLP:conf/emnlp/ShaoGSHDC23,DBLP:journals/corr/abs-2308-11761}. These methods rely on a combination of retrieval techniques such as DPR and LLMs to generate high-quality, context-sensitive results.
Scientific idea proposal can be seen as a special case of retrieval-augmented generation. However, it differs from existing RAG approaches in some key aspects. For example, most RAG algorithms perform retrieval based on relevance, where the query is matched to the keys in the database. However, for idea generation, the number of documents relevant to a specific domain background is often excessively large, making it prone to introducing redundant literature.

\begin{figure*}[t]
\centering
  \includegraphics[width=0.83\textwidth]{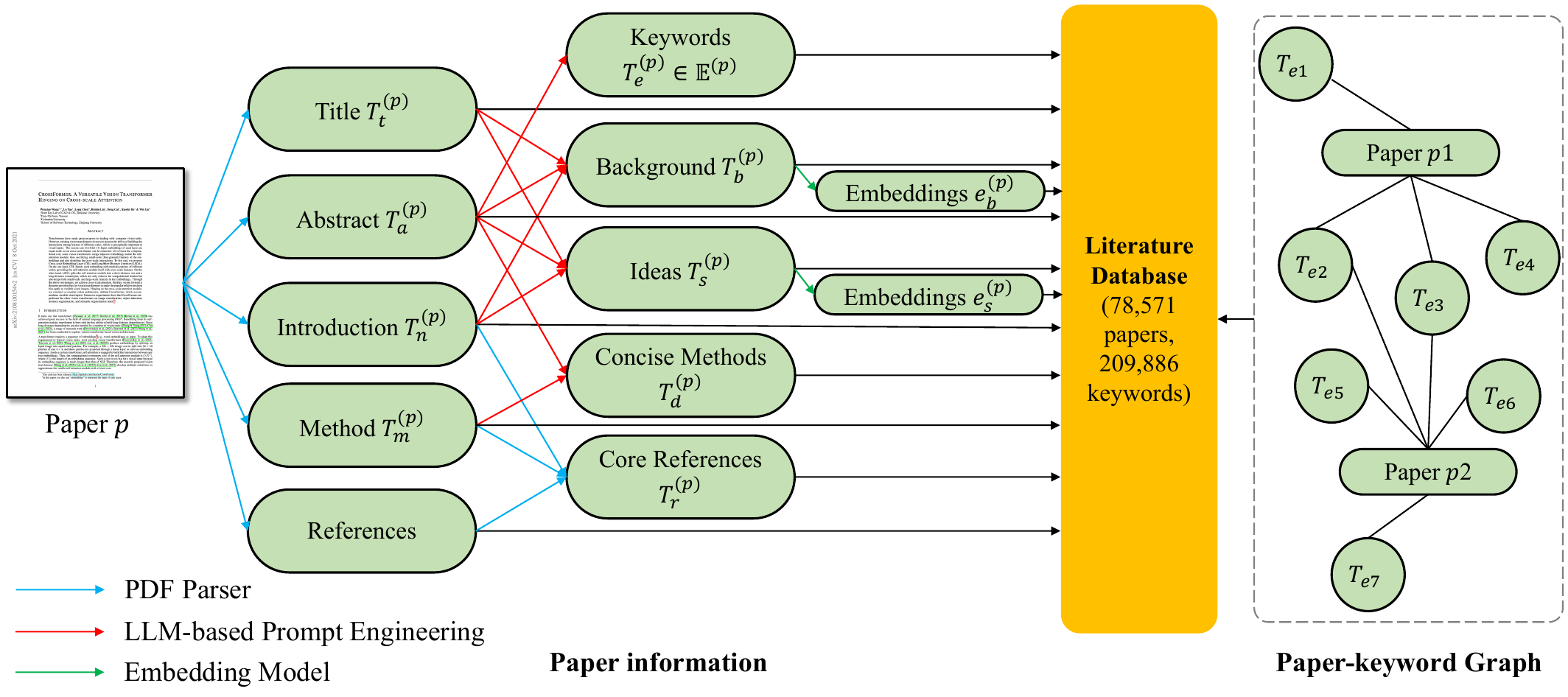}
\caption{The pipeline of constructing the literature database. Paper sections are extracted via a PDF parser, summarized by an LLM, encoded, and stored in the database. A paper-keyword graph linking each paper to its keywords is also stored.} \label{fig:cld}
\end{figure*}

\vspace{-3mm}
\paragraph{Text Embedding Models.} Text embedding models~\citealp{DBLP:conf/emnlp/ReimersG19,DBLP:journals/corr/abs-2212-03533} are designed to represent textual data as high-dimensional vectors, capturing their semantic meaning in a machine-readable format. These embeddings are widely used in various tasks, including text classification, information retrieval, and text matching. Different tasks often require tailored embedding strategies to achieve optimal performance.
For instance, embeddings for information retrieval~\citep{DBLP:journals/corr/abs-2405-17428,DBLP:journals/corr/abs-2409-10173,DBLP:journals/corr/abs-2402-03216,DBLP:conf/emnlp/ZhangZLXDTLYXHZ24,DBLP:journals/corr/abs-2201-10005} are typically asymmetric, meaning that the query and the text being retrieved are encoded using different models or prompts. In this case, a high similarity between the query and the retrieved passage indicates strong relevance. On the other hand, embeddings~\citep{DBLP:conf/emnlp/ReimersG19,DBLP:journals/corr/abs-2409-10173,DBLP:journals/corr/abs-2201-10005} for text matching are symmetric, where both text segments are encoded in the same manner. The similarity score reflects the semantic equivalence between the two text segments.

In addition, we also investigate off-the-shelf literature retrieval tools. These details are provided in Appendix~\ref{sec:lrt} due to space limitations.

\section{Methods}
We introduce a scientific idea proposer called SciPIP, which comprises two key phases: literature retrieval and idea generation. Since existing retrieval algorithms and literature databases are not directly applicable to SciPIP, we first construct a database in Section~\ref{sec:liter-data-constrct}. This process involves summarizing the key information from each paper and encoding these summaries into embeddings. Subsequently, we elaborate on our multi-granularity literature retrieval approach in Section~\ref{sec:liter-retrive}. Finally, inspired by the way humans approach research, we present a dual-path idea generation framework in Section~\ref{sec:idea-proposal}, which leverages the retrieved literature and the internal knowledge of large language models (LLMs) to propose innovative and feasible ideas.

\begin{figure*}[t]
\centering
  \includegraphics[width=0.84\textwidth]{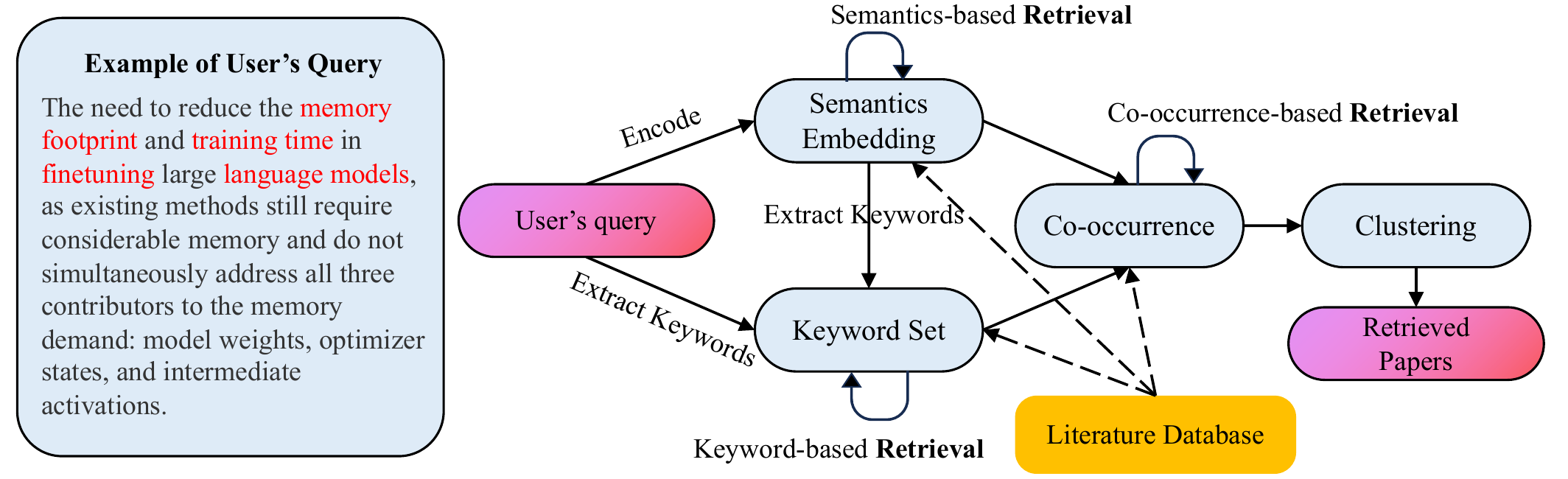}
\caption{The pipeline of SKC-based literature retrieval and literature clustering. Red words in the user's query are entity examples.} \label{fig:figure_retrieval}
\end{figure*}

\subsection{Literature Database Construction} \label{sec:liter-data-constrct}


General literature retrieval either relies on keyword matching or encodes sections of papers or fixed-length chunks to build an index. However, the former overlooks semantic information, and the latter introduces significant noise.
We argue that for scientific retrieval aimed at supporting idea proposal, the focus should be on the background, including the problems or challenges within a research field, as well as the ideas or methods presented in the target literature.
Therefore, by summarizing the literature in advance and extracting the most relevant information for idea proposal to build the index, the effectiveness of literature retrieval can be significantly improved.

To this end, as illustrated in Figure 1, we compile a dataset using papers published in the past decade from several top-tier academic conferences. A PDF parser is then employed to extract titles, abstracts, introductions, methods, and references.
Considering that the background of a paper is typically described in the title, abstract, and introduction, we use these sections as input and design a prompt template. By leveraging prompt engineering, we extract the background and ideas information effectively.
Similarly, another two prompt templates are developed to summarize the keywords and concise methods of the paper. Formally, 
\begin{equation}
\begin{aligned}
(T_b^{(p)}, T_s^{(p)}) &= f(\tau_1, T_t^{(p)}, T_a^{(p)}, T_n^{(p)}), \\ 
\mathbb{E}^{(p)} &= f(\tau_2, T_a^{(p)}), \\
T_d^{(p)} &= f(\tau_3, T_m^{(p)}),
\end{aligned}
\end{equation}
where $f$, $T_t^{(p)}, T_a^{(p)}, T_n^{(p)}, T_m^{(p)}$ are the LLM, the paper $p$'s title, abstract, introduction, and method sections, respectively. $\mathbb{E}^{(p)}, T_b^{(p)}, T_s^{(p)}, T_d^{(p)}, T_r^{(p)}$ are extracted keywords, background, ideas, concise methods, and core references, as shown in Figure~\ref{fig:cld}. $\tau_i, i \in \{1, 2, 3\}$ represent our designed prompt templates, and details of these prompt templates are provided in Table~\ref{tab:ee}.
After summarization, we encode the extracted background and ideas into embeddings using a text-matching embedding model and construct the retrieval index, as shown in Equation~\eqref{equ:embed}.
\begin{equation}
\begin{aligned} \label{equ:embed}
e_b^{(p)} = g(T_b^{(p)}), e_s^{(p)} = g(T_s^{(p)}),
\end{aligned}
\end{equation}
where $e_*^{(p)}$ represents embeddings, which are generated offline and stored in the database, ensuring efficient retrieval.

Besides, to enhance retrieval efficiency, we construct a paper-keyword graph within the database, capturing the relationships between papers and keywords. Specifically, we record all co-occurrence relationships of papers and keywords. As illustrated in Figure~\ref{fig:cld}, if a keyword $T_{e1}$ is mentioned in paper $p1$, an edge is created between the corresponding paper and keyword nodes. This structure facilitates efficient graph-based traversal and retrieval.

\subsection{Multi-granularity Literature Retrieval}  \label{sec:liter-retrive}

Based on our constructed database, we present a multi-granularity literature retrieval framework that integrates three dimensions: keywords, semantics, and citation relationships. Notably, SciPIP does not simply aggregate the retrieval results from keywords, semantics, and citations. Instead, it introduces SKC-based (meaning semantics, keyword, and co-occurrence) retrieval approach that allows the three dimensions to leverage information from one another, thereby enhancing the overall retrieval performance, as illustrated in Figure~\ref{fig:figure_retrieval}.
Furthermore, to mitigate redundancy in the retrieval results, we apply a clustering method to group papers with similar ideas, ensuring that the final retrieval results are free from highly similar references.

\paragraph{Semantics-based Retrieval.} 
As shown in Figure~\ref{fig:figure_retrieval}, during the retrieval process, the user's query $T_b^{(u)}$ generally includes the background of a research field. We first encode the query as $e_b^{(u)}$ using the same embedding model as constructing the database and then compare it with the background embeddings of the papers stored in our database. This approach enables the retrieval of literature that addresses similar challenges, providing more targeted and relevant results.
Formally, let this subset of retrieved papers be denoted as $\mathbb{N}_1$,
\begin{equation}
\label{equ:n1}
    \mathbb{N}_1 = \{p | e_b^{(p)} \in \text{TopK}(cosine(e_b^{(u)}, e_b^{(i)})) \text{ for } i \in \mathbb{D} \},
\end{equation}
where $p$ or $i$ represents a paper in the literature database.

\paragraph{Keyword-based Retrieval.}
Semantic-based retrieval returns a set of papers ($\mathbb{N}_1$) whose backgrounds align closely with the user's query. Entities associated with these papers can play a crucial role in inspiring new ideas. To leverage this, as shown in Figure~\ref{fig:figure_retrieval}, we extract entities from both the user's query and the results of semantic-based retrieval. Specifically, keywords in the user's query are extracted with prompt engineering (The prompt used for this task is detailed in the Appendix~\ref{sec:pt}.) with an LLM, while keywords for papers in $\mathbb{N}_1$ are retrieved directly from the literature database.

For a more thorough retrieval, we further expand the keyword set by employing a keyword-neighborhood-based approach. That is, for an keyword $T_e$ in the current keyword set $\mathbb{E}_1$, any paper $p$ that includes $T_e$ will have its other keywords included in the candidate keyword set. While this may introduce redundant or even noisy keywords, we also employ a keyword filtering method and details are included in Appendix~\ref{sec:kf}.

The final expanded keyword set is denoted as $\mathbb{E}^{(u)}$. Keywords represent concepts that are most relevant to a paper's topic. A paper is likely to be useful if it contains keywords matching those in $\mathbb{E}^{(u)}$. Consequently, for any keyword $T_e$ in the set $\mathbb{E}^{(u)}$, as shown in Equation~\eqref{eq:n2}, we search the database for papers that also contain $T_e$, marking all retrieved papers as the set $\mathbb{N}_2$.

\begin{equation} \label{eq:n2}
    \mathbb{N}_2 = \{p | \exists T_e \in \mathbb{E}^{(u)} \wedge T_e \in T^{(p)}_b, p \in \mathbb{D}\}.
\end{equation}

\paragraph{Co-occurrence-based Retrieval.} 

In practice, we often encounter pairs of papers, $p_1$ and $p_2$, which are neither similar in terms of keywords nor semantic themes but are frequently cited together. This co-citation suggests that researchers have identified a latent relationship between $p_1$ and $p_2$ in prior studies.
To capture and leverage such implicit connections, we propose a literature retrieval method based on citation co-occurrence. As illustrated in Figure~\ref{fig:figure_retrieval}, for any paper $p_1$ that has already been retrieved, if $p_2$ is frequently cited alongside $p_1$ in other papers, $p_2$ will be included in our expanded literature retrieval set. Formally,
\begin{equation}
    \mathbb{N}_3 = \{p_2 | p_1 \in (\mathbb{N}_1 \cup \mathbb{N}_2) \wedge \text{co-cite}(p_1, p_2) \},
\end{equation}
where co-cite means $p_1$ and $p_2$ are often simultaneously cited by other papers. In practice, we select the $nCOCITE$ papers that are most frequently co-cited with each paper.

Finally, the whole retrieved papers can be represented as $\mathbb{N} = \mathbb{N}_1 \cup \mathbb{N}_2 \cup \mathbb{N}_3$.

\paragraph{Literature Clustering and Filtering.} The SKC framework can retrieve hundreds of papers; however, not all of them will be practically referenced. For instance, when describing the background of a research field in a paper, researchers typically cite only a few representative papers rather than exhaustively referencing all relevant literature. To emulate this practice, we propose an idea-similarity-based clustering method.

Initially, each retrieved paper is treated as an individual cluster. Then, for any two clusters $\mathbb{A}$ and $\mathbb{B}$, they will be merged if the summary similarity between any pair of papers from $\mathbb{A}$ and $\mathbb{B}$ exceeds a predefined threshold $\tau$. Here, the similarity between two papers is measured by the cosine similarity of their summary embeddings (\ie, summary in Figure~\ref{fig:cld}).
After clustering, we set a total number of retrieved papers to $nPapers$ and evenly select the same number of papers from each cluster. This approach ensures diversity while reducing redundancy in the retrieved literature.

\begin{figure}[t]
\centering
  \includegraphics[width=0.4\textwidth]{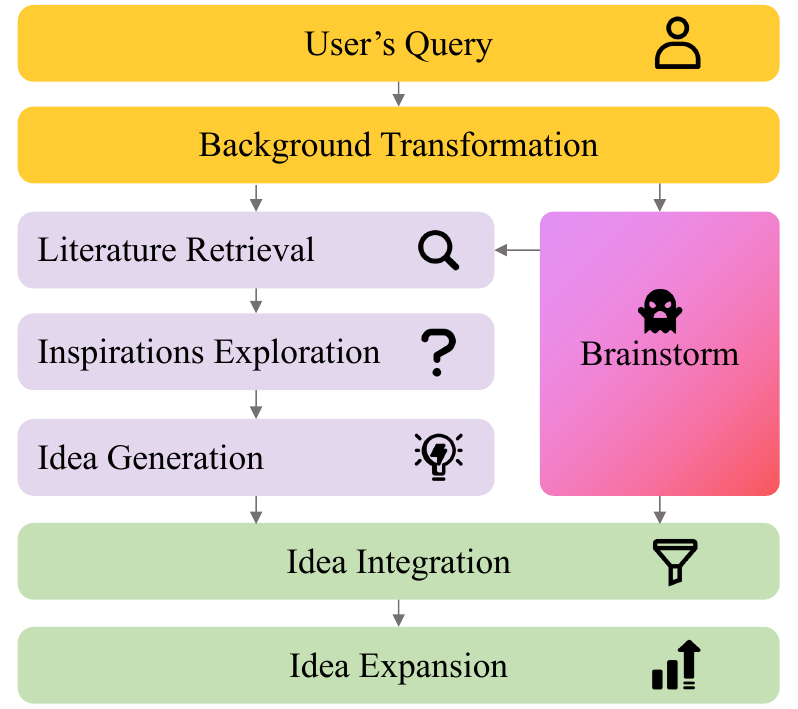}
\caption{The pipeline of SciPIP for idea proposal.}
\label{fig:idea_proposal}
\end{figure}

\subsection{Dual-path Idea Generation}  \label{sec:idea-proposal}

Researchers typically conceive new ideas through two distinct cognitive processes. On one hand, they engage in a thorough review of relevant literature, drawing inspiration from existing works to formulate novel ideas. On the other hand, they rely exclusively on their internal knowledge to brainstorm innovative concepts independently.

Inspired by these cognitive processes, we propose a dual-path framework for LLM-based idea generation that emulates the reasoning patterns of researchers, as illustrated in Figure~\ref{fig:idea_proposal}. Given a user's query, the first path utilizes the literature retrieved in Section~\ref{sec:liter-retrive} to extract insights that may address the user's challenges, subsequently generating ideas based on these insights. In contrast, the second path generates ideas without referencing external literature, relying entirely on the LLM's internal knowledge. Finally, the ideas generated by both paths are integrated and refined, yielding a comprehensive and polished set of outputs.

\subsubsection{Brainstorm-Based Generation}

This pathway generates ideas by relying exclusively on the LLM's internal knowledge. However, we observe that the generation process is highly sensitive to the formulation of the user's query. Specifically, the LLM may misinterpret the query if it is overly brief. To ensure more accurate brainstorming, we first transform the user's query into a more detailed and structured format through prompt engineering. The LLM is instructed to carefully explain each technical term and supplement the query with additional context regarding why the stated problem constitutes a significant challenge. The specific prompt used for this transformation is provided in Appendix~\ref{sec:pt}. This process ensures that the transformed background explicitly includes a thorough description and a complete understanding of the query.

During the brainstorming phase, the LLM is provided with the transformed background and tasked with generating 3 to 4 ideas, each described in several sentences. The prompt used for this brainstorming process is detailed in Appendix~\ref{sec:pt}.

Following the brainstorming phase, we extract keywords from the generated ideas through prompt engineering. These keywords are then merged with those extracted from the background, providing a more robust foundation for subsequent literature retrieval and enhancing the overall thoroughness of the process.

\subsubsection{Retrieval Based generation}

This pathway generates ideas by leveraging insights from retrieved literature. A critical challenge in this phase lies in effectively incorporating the retrieved literature into large language models (LLMs). Existing approaches often utilize titles or abstracts of retrieved papers as input. However, these components, along with the introduction sections, frequently lack the granularity required to comprehensively describe the underlying methodologies. Consequently, the insights derived by LLMs from such inputs are inherently limited. An alternative strategy involves inputting the entire paper to provide richer contextual information. Nevertheless, this approach introduces significant challenges, particularly in managing extremely long contexts, especially when a substantial number of papers are retrieved.

To address these limitations, we employ an LLM to summarize the methodology sections of the retrieved papers, aiming to generate concise yet informative descriptions. These descriptions are designed to be more detailed than the introduction but more succinct than the full methodology section, while excluding redundant content such as background information typically found in introductions. Empirical observations indicate that zero-shot instructions often fail to capture the core aspects of the methodology sections. To overcome this, we adopt a few-shot learning approach, where manually curated summaries of methodology sections are provided as demonstrations to guide the LLM in producing more accurate and focused summaries. The specific prompts used in this process are detailed in Appendix~\ref{sec:pt}.

Subsequently, the concise methodology summaries and the transformed background information are used as inputs to prompt the LLM to extract relevant inspirations. These inspirations focus on how the retrieved papers can address the challenges outlined in the transformed background. The extracted inspirations are further condensed, making them suitable as inputs for the idea generation phase. If the LLM determines that a particular paper offers no valuable insights, it is excluded from further consideration. Finally, the transformed background and the extracted inspirations are combined and fed into the LLM, which is then prompted to synthesize these insights and generate novel ideas. The specific prompts employed in this step are also provided in Appendix~\ref{sec:pt}.

\paragraph{Idea Integration and Expansion.} Through an explicit dual-path generation process, two distinct groups of ideas are produced. These ideas are subsequently provided to the LLM for integration into approximately five coherent ideas. However, the integrated ideas are often brief and lack sufficient clarity. To address this, we employ prompt engineering to expand these ideas into more detailed versions. During this process, we observed that the quality, style, and level of detail in the expanded ideas are challenging to control solely through instructions. Therefore, we retrieve a ``concise method'' section from an existing paper as a demonstration and prompt the LLM to expand the proposed ideas accordingly. The prompts used for integration and expansion are detailed in Appendix~\ref{sec:pt}.

\begin{table}[t]\footnotesize
\centering  
\caption{The subjective evaluation metrics of ideas quality. Higher means better.}  
\label{tab:metrics}  
\setlength{\tabcolsep}{1mm}{
    \resizebox{0.49\textwidth}{!}{
\begin{tabular}{l | m{0.26\textwidth} | c}  
    \toprule  
        Metric & Definition & Rating \\
    \midrule
        Novelty & The idea is novel. & 0 \textasciitilde 5 \\
        Clarity & The idea is clearly explained.& 0 \textasciitilde 5 \\
        Feasibility & The idea will work (be effective) in my experience. & 0 \textasciitilde 5 \\
        Relevance & The idea is relevant to the research background. & 0 \textasciitilde 5 \\
        Helpfulness & The idea is helpful to researchers. & Yes / No \\
    \bottomrule
\end{tabular}
}
}
\end{table}

\begin{table*}[t]
\centering
\caption{The percentage of novelty, clarity, feasibility, and relevance ratings $\ge 4$ of proposed ideas. ``\#'' means ``the number of'', and ``bgs.'' is short for backgrounds. $^\dagger$ means the proposed ideas are drawn from the paper's official code repository. $^\ddag$ means the running code is generated.}
\label{tab:exp:ss}
    \setlength{\tabcolsep}{1mm}{
    \resizebox{0.8\textwidth}{!}{
        \begin{tabular}{l|l|cr|rrrr|r}  
            \toprule  
            Domain & Methods & \#Bgs. \& \#Ideas & LLM & Novelty & Clarity & Feasibility & Relevance & Helpful \\
            \midrule
            \multirow{5}{*}{NLP} & GPT-4o & 34 / 103 & GPT-4o & 16.50\% & 16.50\% & 18.45\% & 64.08\% & 38.83\% \\
            & AI Scientist$^\dagger$ & 3 / 30 & GPT-4o & 20.00\% & (100\%)$^\ddag$ & 23.33\% & - & 46.67\% \\
            & SciPIP (Brainstorm) & 34 / 123 & GPT-4o & 26.02\% & 34.15\% & 24.39\% & 62.60\% & 51.22\% \\
            & SciPIP (Retrieval) & 34 / 167 & GPT-4o & 22.16\% & 31.74\% & 35.33\% & \textbf{68.26\%} & 49.70\% \\
            & SciPIP & 34 / 169 & GPT-4o & \textbf{27.22\%} & \textbf{36.69\%} & \textbf{38.46\%} & 66.86\% & \textbf{53.25\%} \\
            \midrule
            \multirow{4}{*}{CV} & GPT-4o & 25 / 92 & GPT-4o & 13.04\% & 16.30\% & 14.13\% & 42.39\% & 47.83\% \\
            & SciPIP (Brainstorm) & 25 / 97 & GPT-4o-mini & 20.62\% & 26.80\% & 24.74\% & \textbf{47.42\%} & 72.16\% \\
            & SciPIP (Retrieval) & 25 / 115 & GPT-4o-mini & 18.26\% & 18.26\% & \textbf{26.96\%} & 43.48\% & 67.83\% \\
            & SciPIP & 25 / 116 & GPT-4o-mini & \textbf{22.41\%} & \textbf{30.17\%} & \textbf{26.72\%} & \textbf{46.55\%} & \textbf{74.14\%} \\
            \bottomrule
        \end{tabular}
        }
    }
\end{table*}

\section{Experiments}

\subsection{Implementation Details}

\paragraph{Literature Database.} We curate a comprehensive dataset comprising 78,571 papers from nine prominent conferences in machine learning, natural language processing, and computer vision (ICML, NeurIPS, ICLR, ACL, EMNLP, NAACL, CVPR, ICCV, and ECCV) published over the past decade. The backgrounds, ideas, and other relevant information from these papers were summarized using GPT-4o. The summarized content was subsequently encoded using jina-embedding-v3~\cite{DBLP:journals/corr/abs-2409-10173} to facilitate efficient retrieval and analysis.
\vspace{-2mm}
\paragraph{Literature Retrieval.} For the retrieval process, we empirically set the hyperparameters as follows: $TopK=5$, meaning that the semantics-based retrieval returns the top 5 most relevant papers, and $nCOCITE=1$, indicating that co-occurrence retrieval adds one additional paper for each initially retrieved paper. The similarity threshold $\tau$ is set to 0.8, and $nPapers$ is fixed at 10 based on empirical validation. An ablation study is conducted to evaluate the impact of this hyper-parameter on the overall performance.
\vspace{-2mm}
\paragraph{Test Data Construction.} To address the challenges associated with manually constructing user's query, we adopt a systematic approach by randomly selecting 34 papers from ACL 2024 and 25 papers from CVPR 2024. The research backgrounds from these papers are directly utilized as test cases. Importantly, the selected papers had not been pre-released on arXiv prior to their official publication, thereby mitigating the risk of data contamination\footnote{The LLMs employed for idea generation (GPT-4o and GPT-4o-mini) have a data cutoff of October 2023.}.
\vspace{-2mm}
\paragraph{Idea Proposals Evaluation.} The evaluation of proposed ideas is conducted by a panel of 12 human researchers. They are graduate students or PhD candidates in the field of artificial intelligence, each with expertise in natural language processing or computer vision. All evaluators had published at least one first-author paper at top-tier academic conferences, ensuring a high level of domain expertise. They are tasked with assessing each proposed idea across five dimensions: novelty, clarity, feasibility, relevance, and helpfulness, as detailed in Table~\ref{tab:metrics}. Additionally, a pairwise comparison experiment was conducted, wherein evaluators were presented with two ideas generated under the same background and asked to select the one that performed better in terms of novelty, clarity, and other specified criteria.



\begin{figure}[t]
\centering
  \includegraphics[width=0.4\textwidth]{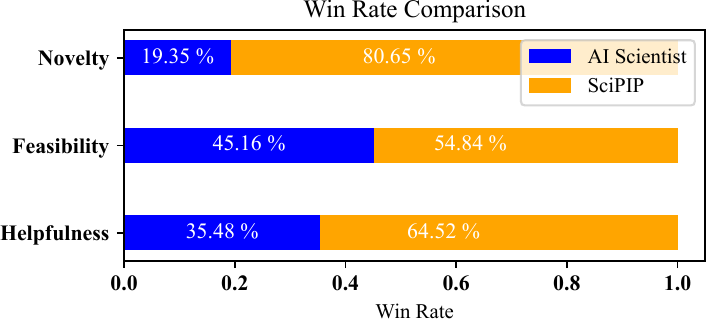}
\caption{Comparison between SciPIP and AI Scientist. Both frameworks takes 3 backgrounds as input and generate about 30 ideas. GPT-4o-mini is used during generation.} \label{fig:winrate}
\end{figure}
\vspace{-3mm}
\subsection{Results and Analysis}

\paragraph{Independent Ratings of Ideas.} 

Through the construction of research backgrounds in both NLP and CV domains, we conduct a comprehensive comparison of proposal quality among SciPIP, GPT-4o, and AI Scientist. In this comparison, GPT-4o generates ideas directly through prompt engineering, utilizing the user's query as input. For AI Scientist, its ideas are either drawn from or generated using its official code repository, resulting in the use of different backgrounds compared to our approach.

All generated ideas were evaluated by invited researchers, and we calculated the percentage of ideas that received favorable ratings ($\ge$ 4) from the evaluators. The results are presented in Table~\ref{tab:exp:ss}. As demonstrated, all three versions of SciPIP outperform GPT-4o and AI Scientist. Notably, SciPIP (Brainstorm) leverages internal knowledge similar to AI Scientist but produces ideas with significantly higher novelty. We hypothesize that this is because AI Scientist's input includes an existing paper and codebase, which may constrain its creative scope. The clarity of AI Scientist's ideas is consistently rated highly, as it generates executable code that provides clear explanations of the proposed ideas. Furthermore, the results reveal that brainstorming tends to generate more novel ideas compared to literature retrieval-based generation (25.77\% vs. 22.14\%), though the former often lacks sufficient feasibility (24.14\% vs. 35.33\%). SciPIP integrates the outputs of both brainstorming and retrieval-based generation, achieving an improved balance between novelty and feasibility while yielding improved results.

SciPIP also demonstrates superior performance in the CV domain, highlighting its generalizability across fields. However, its performance in CV is slightly inferior to that in NLP, particularly in terms of clarity. We attribute this discrepancy to the relatively weaker performance of GPT-4o-mini compared to GPT-4o. An intriguing observation is that 74.17\% of CV ideas are rated as helpful, surpassing the corresponding percentage for NLP ideas (53.14\%). This discrepancy may stem from inconsistent evaluation standards, as the evaluation tasks were distributed according to the volunteers' research fields, resulting in CV and NLP ideas being assessed by two distinct groups of researchers.
\vspace{-4mm}
\paragraph{Win Rates between SciPIP vs. AI Scientist.} We conducted a comparative analysis of the win rates between SciPIP and AI Scientist. AI Scientist is designed to generate novel ideas by leveraging existing papers, code bases, and experimental data as input, without requiring predefined challenges or problems. To ensure a fair comparison under identical conditions, we extracted the proposed ideas from AI Scientist and utilized an LLM to infer the underlying problems they aimed to address. These inferred backgrounds were then used as input for SciPIP, enabling a direct comparison of the ideas generated by both systems within the same contextual framework. The comparative win rates are illustrated in Figure~\ref{fig:winrate}. The results demonstrate that SciPIP consistently outperforms AI Scientist across all evaluation metrics. Notably, human researchers rated 80\% of the ideas generated by SciPIP as more novel compared to those produced by AI Scientist. Additionally, SciPIP exhibits superior performance in terms of feasibility and helpfulness, further underscoring its effectiveness in generating high-quality research ideas.
\vspace{-2mm}

\begin{figure}[t]
\centering
  \includegraphics[width=0.48\textwidth]{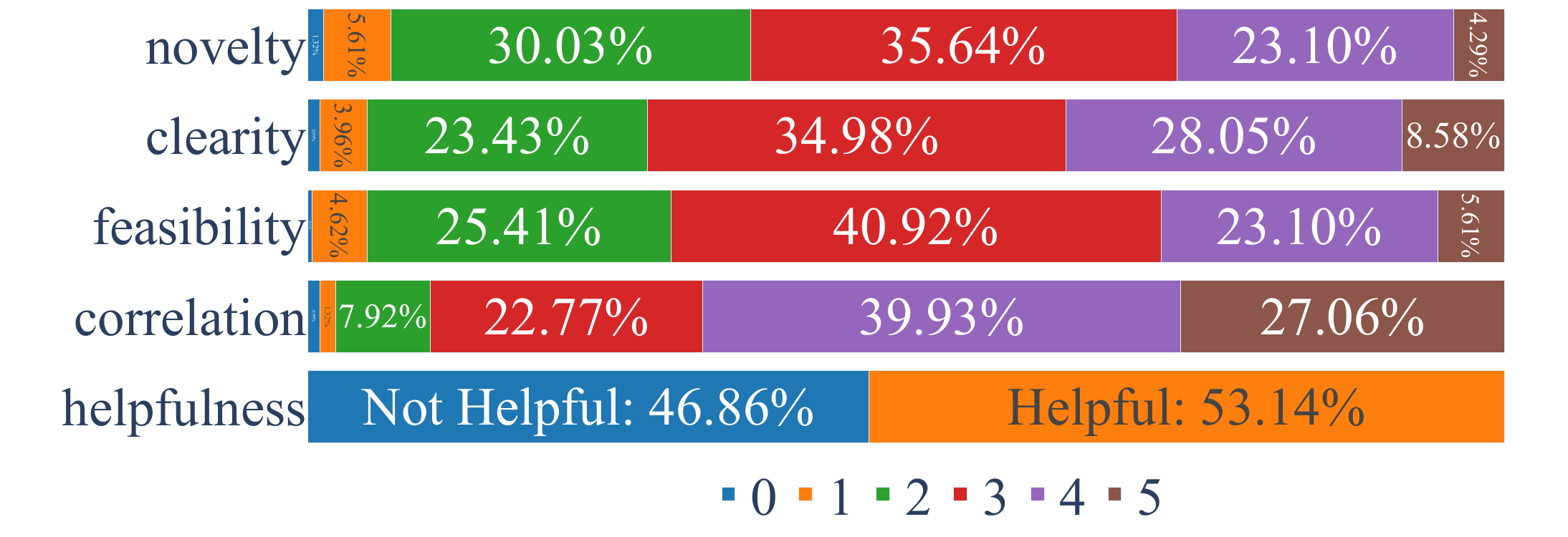}
  \vspace{-8mm}
\caption{The distribution of human ratings of SciPIP proposed NLP ideas. GPT-4o is used during generation.} \label{fig:figure_distribution}
\end{figure}

\begin{table*}[t]\scriptsize
	\centering  
	\caption{Demonstration of a SciPIP proposed idea. Only the concise version is provided due to space limitation, and the detailed version is in the Appendix~\ref{sec:spi}.}  
	\label{tab:demo}  
    \resizebox{0.95\textwidth}{!}{
	\begin{tabular}{l | m{0.75\textwidth}}  
		\toprule  
            \textbf{Background} & 1. The need to overcome the limitations of traditional RAG systems, which struggle with processing lengthy documents and filtering out irrelevant content, leading to inaccurate response generation.
            
2. The desire to eliminate the suboptimal chunking process, which disrupts semantic coherence and results in incomplete and incoherent retrieved information. \\
            \midrule
            \textbf{Proposed Idea} & \textbf{Multi-Stage Contextual Thinning with Semantic Twin Projections}
            
Proposed by integrating Dynamic Attention-Based Document Thinning and Semantic Twin Networks, introduce a multi-stage thinning process that refines document content leveraging semantic projections. Initially, a broad stroke thinning reduces document length by filtering with dual embeddings while maintaining high-level context. Subsequently, targeted semantic twin projections help identify and concentrate on key chunks according to their contextual relevance. This dual-phase smart thinning reduces the need for arbitrary chunking, preserving semantic connections and enhancing the efficiency and quality of generated outputs by guiding the generative model to produce well-informed, coherent responses.
\\
\midrule
\textbf{Detailed Idea} & Please refer to Appendix~\ref{sec:spi} for details. \\
		\bottomrule
	\end{tabular}
    }
\end{table*}

\begin{figure}[t]
\centering
  \includegraphics[width=0.38\textwidth]{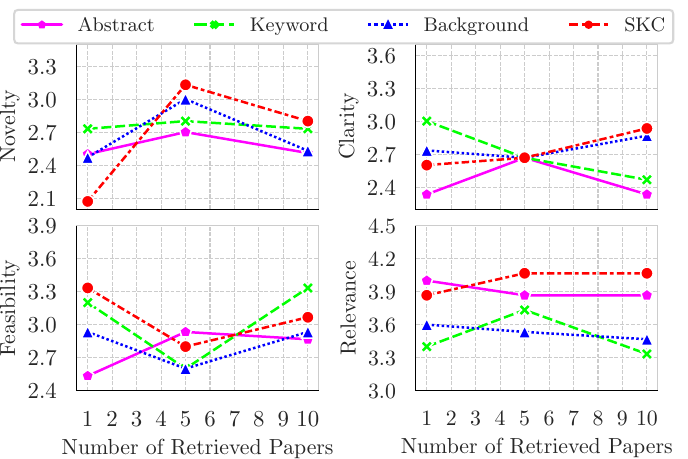}
\caption{Ablation studies on different literature retrieval methods and the number retrieved papers. Y-axis means the average ratings over all ideas.} \label{fig:ablation}
\end{figure}
\vspace{-2mm}

\paragraph{Distribution of Ratings.} The distribution of human ratings for NLP ideas generated by SciPIP is presented in Figure~\ref{fig:figure_distribution}. The results demonstrate that ratings $\ge 3$ constitute the majority across the novelty, clarity, and feasibility metrics. Given that the rating scale is divided into six segments, a score of $3$ represents a relatively positive evaluation. Furthermore, over 66\% of the ideas achieve relevance scores $\ge 4$, indicating that SciPIP demonstrates strong capability in generating ideas that are well-aligned with the provided background. Additionally, approximately 53.14\% of the ideas are rated as helpful, suggesting that more than half of the generated ideas are considered valuable by human evaluators.


\vspace{-2mm}
\paragraph{Ablations on Retrieval Methods.} We perform ablation studies to evaluate the impact of various literature retrieval methods and the number of retrieved papers on the overall performance. GPT-4o-mini is employed for idea generation in these experiments. We do not use larger models like GPT-4o because of the efficiency concern. The corresponding results are presented in Figure~\ref{fig:ablation}. The terms ``Abstract'' and ``Background'' refer to the encoding of the abstract section or our extracted background information as the retrieval vector, followed by vector-based retrieval. ``Keyword'' denotes the application of keyword-based retrieval exclusively. ``SKC'' represents our proposed multi-granularity retrieval method. As demonstrated in the results, SKC achieves the most balanced and superior performance across the evaluated metrics. No significant correlation is observed between the number of retrieved papers and the ratings. Therefore, to maintain a balance among the four metrics, we consistently retrieve 10 papers in subsequent experiments.
\vspace{-5mm}
\paragraph{Case Study.} Among the 169 ideas generated, as summarized in Table~\ref{tab:exp:ss}, several achieved high evaluation scores. One such idea, rated 4 for both novelty and feasibility, is presented in Table~\ref{tab:demo}, while additional examples are included in the supplementary materials. The highlighted idea addresses a key limitation of existing retrieval-augmented generation (RAG) systems, which often rely on simplistic chunking methods. The proposed idea introduces a representation-based importance metric and semantic twin networks to selectively extract key information from lengthy documents. This approach is expected to enhance the performance of current RAG algorithms.
\vspace{-2mm}
\section{Conclusions and Limitations}
In this paper, we propose a novel method, SciPIP, for generating scientific paper ideas and demonstrate its effectiveness in the domains of natural language processing and computer vision. Experimental results indicate that SciPIP leverages the capabilities of large language models (LLMs) to generate numerous innovative ideas. These ideas exhibit strong potential in terms of novelty, feasibility, clarity, and other critical dimensions. However, despite these promising results, the study has certain limitations. Specifically, the feasibility of the generated ideas was evaluated subjectively based on humans. Moreover, this study focuses on two specific fields including NLP and CV, and we intend to extend SciPIP to other fields and disciplines to assess and demonstrate its broader applicability and generalizability.

\bibliography{example_paper}
\bibliographystyle{icml2025}

\newpage
\appendix
\onecolumn
\section{Related Works About Literature Retrieval Tools} \label{sec:lrt}
Literature retrieval is a fundamental process in scientific research, and several well-established scientific literature search engines are widely used. Among them, Google Scholar and Semantic Scholar are two of the most popular platforms, offering literature retrieval services across all research fields. However, these platforms do not provide access to the full text of the indexed papers. In contrast, arXiv maintains its own comprehensive literature database, which not only facilitates literature retrieval but also provides full-text access. ArXiv primarily focuses on fields such as mathematics, physics, and computer science. Similarly, PubMed and PMC (PubMed Central) cater specifically to the field of medicine, offering specialized literature retrieval services. Despite their usefulness, these databases predominantly rely on keyword matching for retrieval, which often leads to imprecise results. Consequently, users are left with the burden of manually filtering out irrelevant literature to find the information they need.

\section{SciPIP Proposed Ideas.} \label{sec:spi}

The detailed idea of demonstration shown in Table~\ref{tab:demo} is given as follows. \textcolor{red}{All the following contents in Appendix~\ref{sec:spi} are generated by SciPIP.} Other demonstrations of SciPIP proposed ideas are provided in the supplementary materials.

\subsection{Detailed Idea}

To improve Retrieval-Augmented Generation (RAG) systems' handling of lengthy documents, the proposed ``Multi-Stage Contextual Thinning with Semantic Twin Projections'' algorithm systematically reduces document content through a novel dual-phase process that enhances semantic integrity and contextual relevance.

\subsubsection{Overview of the Multi-Stage Contextual Thinning Process}

The process consists of two main phases: Dynamic Attention-Based Document Thinning and Semantic Twin Projections. Each phase progressively refines the document, enhancing relevance and coherence before it is processed by the generative module.

\subsubsection{Phase 1: Dynamic Attention-Based Document Thinning} 

In this initial phase, the algorithm applies attention mechanisms to focus on contextually important parts of a document:

\paragraph{Dual Embedding Representation}
\begin{itemize}
    \item Utilize dual embedding models where $E_{global}$ captures overarching document themes, while $E_{local}$ focuses on granular details.
    \item Compute global attention scores $A_{global}$ and local attention scores $A_{local}$ using the formula:
$$
A_{global} = \text{softmax}(Q_{global} \cdot K_{global}^T), \quad A_{local} = \text{softmax}(Q_{local} \cdot K_{local}^T)
$$
where $Q$, $K$ are query and key matrices derived from $E_{global}$ and $E_{local}$ embeddings respectively.
\end{itemize}

\paragraph{Broad Stroke Thinning}
\begin{itemize}
\item Combine these attention scores to prioritize sections of the document that maintain a balance between thematic relevance and detailed context:
  $$
  A_{combined} = \alpha A_{global} + (1 - \alpha) A_{local}
  $$
  where $\alpha \in [0, 1]$ is a tunable parameter to balance global and local attentions.

\item Select document segments with the highest $A_{combined}$ scores to create a reduced version of the document, filtering out less relevant material while retaining the core content.
\end{itemize}

\subsubsection{Phase 2: Semantic Twin Projections}

In the second phase, the methodology refines the thinned document using semantic twin networks to further emphasize key information.

\paragraph{Semantic Twin Network Configuration}
\begin{itemize}

\item Establish twin neural networks, $T_{context}$ and $T_{focus}$, where $T_{context}$ maintains contextual integrity while $T_{focus}$ concentrates on semantic salience.
\item  Feed the reduced document from Phase 1 into both networks to obtain semantic projections.
\end{itemize}

\paragraph{Targeted Chunk Identification}
\begin{itemize}

\item Compare outputs from $T_{context}$ and $T_{focus}$ to discern segments that align strongly on both context and focus dimensions.

\item Sections are prioritized based on the similarity of projections:
  $$
  S_{priority} = \text{cosine\_similarity}(T_{context}(d_i), T_{focus}(d_i))
  $$
  where $d_i$ represents a document segment and cosine similarity measures alignment between the two network outputs.
\end{itemize}

\paragraph{Final Document Synthesis}
Construct the final document synthesis by incorporating segments with high $S_{priority}$ scores, ensuring contextual coherence and specificity.

\subsubsection{Integration with RAG Systems}

The refined document content, now presenting coherent and contextually various segments, feeds into the generative module. The minimized document scope reduces computational overhead and enhances semantic coherence in generated outputs.

\subsubsection{Benefits}
\begin{itemize}

\item  Reduced need for arbitrary chunking: By employing dynamic attention-based and semantic projections, the algorithm ensures that document segmentation is guided organically by semantic relevance rather than arbitrary heuristics.
\item  Improved Semantic Coherence**: Preservation of logical connections between segments enhances the quality of generative responses.
- Increased Efficiency: By narrowing down documents to their most relevant components, the generative model requires fewer resources while maintaining response quality.
\end{itemize}

This approach not only addresses the inherent limitations of traditional RAG systems when dealing with lengthy documents but also allows more refined and efficient document processing, ultimately boosting the performance of response generation.

\section{Keyword Filtering} \label{sec:kf}
In our multi-granularity retrieval framework, we expand the keyword set by employing a keyword-neighborhood-based approach. For an keyword $T_e$ in the current keyword set $\mathbb{E}_1$ , any paper $p$ that includes $T_e$ will have its other keywords included in the candidate keyword set.
However, this approach may introduce redundant or even noisy entities due to the following reasons:
\begin{enumerate}
\item Low-Relevance Co-Occurrences: Two entities with low relevance might appear together in a paper due to the specific content requirements of that paper.
\item High-Frequency Terms: Certain high-frequency terms do not effectively characterize a paper. For example, while the user-provided background might include the term ``Transformer'', not all entities co-occurring with ``Transformer'' in other papers are relevant. This is because ``Transformer'' is a commonly used term in many recent publications.
\end{enumerate}
To mitigate these issues, we introduce two filtering mechanisms for neighborhood-based entity expansion:
\begin{enumerate}
\item Co-Occurrence Threshold: An entity is supplemented only if it appears together with another entity in at least $m$ papers.
\item Low-Frequency Selection: Inspired by the TF-IDF algorithm, we hypothesize that entities appearing frequently across the entire literature database are less representative. Thus, we retain only the $n$ entities with the lowest frequency across all literature.
\end{enumerate}

\section{Prompt Templates} \label{sec:pt}

The prompt templates we provide in the appendix are summarized in Table~\ref{apd:tab:summary}.

\begin{table}[btp]
	\centering  
	\caption{Summarization of our used prompts.}  
	\label{apd:tab:summary}  
        \resizebox{\textwidth}{!}{
	\begin{tabular}{l | l}
		\toprule  
            \textbf{Prompts} &  \textbf{Place} \\
            \midrule
            The prompt for summary, background, and main ideas extraction, namely $\tau_1$. & Table~\ref{tab:be} \\
            The prompt for entity extraction, namely $\tau_2$. & Table~\ref{tab:ee} \\
            The prompt for concise methods extraction, namely $\tau_3$. & Table~\ref{tab:ecm} \\
            The prompt for background transformation. & Table~\ref{tab:bt}\\
            The prompt for brainstorming. & Table~\ref{tab:brst} \\
            The prompt for initial idea generation. & Table~\ref{tab:iig} \\
            The prompt for idea integration. & Table~\ref{tab:if} \\
            The prompt for idea expansion. & Table~\ref{tab:ie} \\
		\bottomrule
	\end{tabular}
        }
\end{table}

\begin{table}[t]
	\centering  
	\caption{The prompt for background and ideas extraction, namely $\tau_1$.}  
	\label{tab:be}  
    \resizebox{0.98\textwidth}{!}{
	\begin{tabular}{m{0.2\textwidth} | m{0.75\textwidth}}  
		\toprule  
            \textbf{System Message} & Now you are an expert in extracting key entities from research contents. You are good at identifying the most important keywords or phrases that summarize the main topics or concepts discussed in the content. \\
            \midrule
            \textbf{User Message For Summary} & Task Description:
\newline\linebreak
You are provided with the title, abstract, and introduction of a research paper. Your task is to generate a concise summary of what kind of problem does this paper aim to solve and what methods are proposed to address it. The summary should follow this format:
The problem of [problem] can be addressed by [main idea/approach].
\newline\linebreak
Instructions:
\newline\linebreak
Title: Read the title to understand the general topic of the paper.
Abstract: Read the abstract to get a concise summary of the research, including the problem addressed, the methods used, and the main findings.
Introduction: Read the introduction to gain a deeper understanding of the background, significance, and specific problem the paper addresses, as well as the proposed approach or solution.
Based on the provided information, generate a single sentence that captures the essence of the paper, following the format specified above.
\newline\linebreak
Your Turn:
\newline\linebreak
Given the following paper information:
Title: {title}
Abstract: {abstract}
Introduction: {introduction}
\newline\linebreak
Output:
The problem of [problem] can be addressed by [main idea/approach]. \\
\midrule
\textbf{User Message For Background And Main Ideas} & Please read the title, abstract, and introduction of the paper again, as well as the summary you provided. Complete the following two tasks:

1.Briefly provide the two most critical motivations behind proposing these methods to address the problems. 

2.Briefly provide the three most critical or innovative details of the paper that were not mentioned in your summary (It's best if these details are the new methods or techniques adopted in this paper).
\newline\linebreak
Output:

Motivations:1.[motivation1]. 2.[motivation2]. Details:1.[detail1]. 2.[detail2]. 3.[detail3]. \\
		\bottomrule
	\end{tabular}
    }
\end{table}

\begin{table}[t]
	\centering  
	\caption{The prompt for entity extraction, namely $\tau_2$.}  
	\label{tab:ee}  
	\begin{tabular}{l | m{0.75\textwidth}}  
		\toprule  
            \textbf{System Message} & Now you are an expert in extracting key entities from research contents. You are good at identifying the most important keywords or phrases that summarize the main topics or concepts discussed in the content. \\
            \midrule
            \textbf{User Message} & Task Description:
\newline\linebreak
I will provide you with a content from a research paper. Your task is to extract the key entities from this content. These entities are the most important keywords or phrases that summarize the main topics or concepts discussed in the content.
\newline\linebreak
Instruction:
\newline\linebreak
Content: The content is your key focus, and the extracted entities should be based on the content. In other words, the entities you extract should be concrete manifestations of the main themes and topics discussed in the content.
\newline\linebreak
Your approach should be systematic:

- Start by thoroughly reading the content to understand its main themes and topics.

- Identify and list the key entities that are central to the content.

- Ensure that the entities are relevant, meaningful, and representative of the content.

- Each entity in entities should be no longer than 5 words.

- Each entity in entities should contain at least 2 words.

- The number of entities should be less than or equal to 5.

- Each entity in entities should be nouns or noun phrases.
\newline\linebreak
examples:

\{examples\}
\newline\linebreak
Your turn:

Given the following content:

\{content\}
\newline\linebreak
Your answer should follow this format:

entity1, entity2, entity3, ......\\
		\bottomrule
	\end{tabular}
\end{table}

\begin{table}[t]
	\centering  
	\caption{The prompt for extracting concise methods, namely $\tau_3$.}  
	\label{tab:ecm}  
	\begin{tabular}{l | m{0.75\textwidth}}  
		\toprule  
            \textbf{System Message} & Now you are a researcher in the field of AI with innovative and pioneering abilities. \\
            \midrule
            \textbf{User Message} & \# Task Description:
            
You are an AI researcher conducting studies in a specific domain. Someone has provided you with a methodology section, and your task is to transform it into another style. I will give you an example. The example begins with ``\# Example 1'' and includes a Example Summarized Methods. Then, your task starts with ``\# Your Task'', containing ``Your Methodology Section''. Your job is to transform Your Methodology Section into a Summarized Methods by referring to Example 1. Note that the ideas in Example 1 are unrelated to your idea, so the key focus should be on the style of Example Summarized Methods. You should directly start with your response and do not start with a section title like ``\#\# Your Summarized Methods''.

\# Example 1

\#\# Example Summarized Methods

\{Example Summarized Methods\}

\# Your Task

\#\# Your Methodology Section

\{methodology\}

\#\# Your Summarized Methods
            \\
		\bottomrule
	\end{tabular}
\end{table}

\begin{table}[t]
	\centering  
	\caption{The prompt for background transformation.}  
	\label{tab:bt}  
	\begin{tabular}{l | m{0.75\textwidth}}  
		\toprule  
            \textbf{System Message} & You are a teacher in the field of AI, skilled at clearly explaining AI concepts to students. Your student is an undergraduate in AI with a basic understanding of deep learning. \\
            \midrule
            \textbf{User Message} & \# Task Description:
            
You are teaching your undergraduate about a specific subfield of AI research. You have a brief description of the research background, and now you need to explain its meaning and purpose in detail to your undergraduate. Keep in mind that your undergraduate may be completely unfamiliar with the technical terms in the research background. I will give you an example. The example begins with ``\# Example 1'' and includes a Brief Research Background, several Technical Terms, and the corresponding Detailed Research Background. Then, your task starts with ``\# Your Task'', containing ``Your Brief Research Background'' and ``Your Technical Terms''. Your job is to expand Your Brief Research Background into a Detailed Research Background by referring to Example 1. Note that the research background in Example 1 are unrelated to yours, so the key focus should be on the relationship between the Brief Research Background and the Detailed Research Background. You should directly start with your response and do not start with a section title like ``\#\# Detailed Background''. 

\# Example 1

\#\# Brief Research Background

\{Example user's query\}

\#\# Technical Terms

large language models, kv cache, gpu memory

\#\# Detailed Research Background

\{Example detailed research background\}

\# Your Task

\#\# Your Research Background

\{brief\_background\}

\#\# Your Technical Terms

\{keywords\}
\\
		\bottomrule
	\end{tabular}
\end{table}

\begin{table}[htbp]
	\centering  
	\caption{The prompt for brainstorming.}  
	\label{tab:brst}  
	\begin{tabular}{l | m{0.75\textwidth}}  
		\toprule  
            \textbf{System Message} & Now you are a researcher in the field of AI with innovative and pioneering abilities. You are good at generating creative and original ideas. \\
            \midrule
            \textbf{User Message} & \#\#\# Task Description: 

You are an AI researcher tasked with brainstorming initial, innovative ideas to address a given research problem in AI. Focus on generating diverse and creative approaches rather than finalized methods. The ideas can be rough and in their infancy but should cover a range of possible directions that could be explored further.
\newline\linebreak
\#\#\# Information Provided:

- **Research Background**: \{background\}
\newline\linebreak

\#\#\# Approach:

Your brainstorming should be systematic:

- **Step 1**: Thoroughly understand the research background.

- **Step 2**: Generate a list of 4 to 6 high-level ideas or directions that could potentially solve problems in the given background. Be creative, think outside the box, and avoid merely rephrasing existing methods.
\newline\linebreak
\#\#\# Format for Your Response:

Please present 4 to 6 ideas in the following format:

**Idea 1**: [Brief description of the first idea]

**Idea 2**: [Brief description of the second idea]

... \\
\bottomrule
\end{tabular}
\end{table}

\begin{table}[t]
	\centering  
	\caption{The prompt for initial idea generation.}  
	\label{tab:iig}  
	\begin{tabular}{l | m{0.75\textwidth}}  
		\toprule  
            \textbf{System Message} & Now you are a researcher in the field of AI with innovative and pioneering abilities. You are good at using innovative and original methods to solve cutting-edge problems in the field of AI. \\
            \midrule
            \textbf{User Message} & \# Task Description:
You will be provided with a research problem, along with inspirations extracted from related papers. Your task is to identify and combine these inspirations to propose 3 to 4 different ideas to solve the research problem. The ideas should be innovative and try to avoid using the same inspirations repeatedly. Each idea should starts with ``**Idea **:''.

\# Information Provided:

\#\# Research problem

\{Transformed background\}

\#\# Inspirations

\{Inspirations\}
\\
		\bottomrule
	\end{tabular}
\end{table}

\begin{table}[t]
	\centering  
	\caption{The prompt for idea integration.}  
	\label{tab:if}  
	\begin{tabular}{l | m{0.75\textwidth}}  
		\toprule  
            \textbf{System Message} & Now you are a researcher in the field of AI. You are good at selecting the ideas that meet the requirements. \\
            \midrule
            \textbf{User Message} & \#\#\# Task Description: 
            
You will be provided with some ideas you previously generated, and a research background. Your task is to select 5-6 ideas that best address the problems described in the research background (priority) and ideas that are relatively novel and feasible (secondary).

\#\#\# Information Provided:

1. **Ideas**: These are the ideas you previously generated based on the research background and several related papers.

2. **Research Background**: This document describes specific problems and challenges that need to be addressed.

\#\#\# Approach:

Your approach should be systematic:

- **Step 1**: Analyze the research background to understand the specific problems that need solutions.

- **Step 2**: Critically review the ideas, selecting 5-6 ideas that are most effective in solving the problems in the research background (priority) and that are also relatively novel and feasible (secondary).

\#\#\# Specific Information:

I will provide you with specific information now; please use them according to the instructions above:

1. **Ideas**: \{idea\}

2. **Research Background**: \{background\}

\#\#\# Format for Your Response:

Please ensure that your final ideas include 5-6 entries, whose content has not been modified. Don't generate any explanation and just present the filtered ideas as well as their content in the following format:

**Idea 1**: [The first method idea]  

**Idea 2**: [The second method idea]  

**Idea 3**: [The third method idea]  

...\\
		\bottomrule
	\end{tabular}
\end{table}

\begin{table}[t]
	\centering  
	\caption{The prompt for idea expansion.}  
	\label{tab:ie}  
	\begin{tabular}{l | m{0.75\textwidth}}  
		\toprule  
            \textbf{System Message} & Now you are a researcher in the field of AI with innovative and pioneering abilities. You are good at transforming a brief scientific idea into a concrete algorithm. \\
            \midrule
            \textbf{User Message} & \# Task Description:
You are an AI researcher conducting studies in a specific domain. Someone has provided you with a brief scientific idea, and your task is to transform it into a detailed, feasible, and concrete algorithm. If necessary, you may incorporate formulas to elaborate on the algorithm in the Latex format. I will give you an example. The example begins with ``\# Example 1'' and includes a Brief Scientific Idea and its corresponding Detailed Scientific Idea. Then, your task starts with ``\# Your Task'', containing ``Your Brief Scientific Idea''. Your job is to expand Your Brief Scientific Idea into a Detailed Scientific Idea by referring to Example 1. Note that the ideas in Example 1 are unrelated to your idea, so the key focus should be on the relationship between the Brief Scientific Idea and the Detailed Scientific Idea. You should directly start with your response and do not start with a section title like ``\#\# Detailed Scientific Idea''. 

\# Example 1

\#\# Example Brief Scientific Idea

\{Example brief idea\}

\#\# Example Detailed Scientific Idea

\{Example detailed idea\}

\# Your Task

\#\# Your Research Background

\{background\}

\#\# Your Brief Scientific Idea

\{brief\_idea\} \\
		\bottomrule
	\end{tabular}
\end{table}

\end{document}